\documentclass[11pt]{article}

\usepackage[margin=1in]{geometry}
\usepackage[T1]{fontenc}
\usepackage[utf8]{inputenc}
\usepackage{lmodern}
\usepackage{microtype}
\usepackage{amsmath,amssymb}
\usepackage{booktabs}
\usepackage{longtable}
\usepackage{array}
\newcolumntype{L}[1]{>{\raggedright\arraybackslash}p{#1}}
\usepackage{enumitem}
\usepackage{url}
\usepackage{hyperref}
\usepackage{xcolor}
\usepackage{verbatim}
\usepackage{listings}
\usepackage{etoolbox}

\hypersetup{
  colorlinks=true,
  linkcolor=blue,
  citecolor=blue,
  urlcolor=blue,
  pdftitle={GROUND: Reducing Hallucinations in LLM-Based Enterprise Analytics Through Governed Semantic Definitions},
  pdfauthor={Aravind Sasidharan Pillai}
}

\providecommand{\tightlist}{%
  \setlength{\itemsep}{0pt}\setlength{\parskip}{0pt}}
\setlist[itemize]{leftmargin=*}
\AtBeginEnvironment{longtable}{\small}
\title{GROUND: Reducing Hallucinations in LLM-Based Enterprise Analytics Through Governed Semantic Definitions}
\author{Aravind Sasidharan Pillai$^{*1}$\\[0.25em]
\small $^{*1}$Principal Architect, Data Engineering, Cox Automotive Inc, Foster City, CA, USA}
\date{July 2026}

\begin{document}
\maketitle

\begin{abstract}
Natural-language analytics over enterprise data warehouses is an increasingly important use case for large language models, but production adoption is limited by hallucinated metrics, invalid joins, wrong grain, unsafe data access, and unverified explanations. Existing text-to-SQL approaches ground generation in database schemas or retrieved documentation, whereas enterprise reporting requires grounding in governed business semantics: approved metrics, dimensions, join paths, filters, and row-level security. This paper introduces \textbf{GROUND} (Governed Retrieval Over Unified Normalized Definitions), which constrains LLM-generated analytics to a governed semantic layer --- supplying approved definitions, binding intent to approved metrics and dimensions under grain and join constraints, and validating every query against schema, metric, join, grain, filter, security, and cost rules before execution, retrying on violations and abstaining when a request is unanswerable. In the primary synthetic experiment, against three baselines sharing one underlying model --- direct schema-only text-to-SQL, schema-RAG, and semantic-only grounding --- differences isolate the effect of governance context. GROUND is the only system free of measured hallucinations across all six evaluated categories, while every ungoverned system violates row-level security on many questions; critically, a semantic-only condition with exact metric definitions but no access policy still leaks data across tenant boundaries, isolating governance --- not metric fidelity --- as the property grounding-as-context does not supply. This costs roughly 5$\times$ the tokens and 1.3$\times$ the latency of the direct baseline. We replicate the findings on real U.S. NHTSA vehicle-safety data with independent hand-authored gold, and probe GROUND\textquotesingle s boundary with an adversarial set across four models from three providers, including an open-weight model. GROUND\textquotesingle s \emph{enforced} guarantees (filters, row-level security) hold with zero violations on every model --- the benefit is not model-specific --- while judgment-dependent behaviors (refusing an undefined metric, choosing to clarify) remain fallible even on the strongest model. Governance prevents safety-critical filter and row-level-security violations even for weaker models, although result accuracy still tracks raw model capability.

\end{abstract}


\section*{Core contribution}

GROUND is a governed semantic-retrieval and validation framework that constrains LLM-generated analytics to approved enterprise definitions, reducing schema, metric, join, grain, filter, security, and explanation hallucinations. We contribute the framework, an open benchmark of 100 governed reporting questions with a deterministic, semantic-layer-driven gold-SQL generator, and a controlled four-way comparison showing that, in our benchmarks, row-level security compliance was not achievable from semantic definitions alone --- it required an explicit access policy enforced by validation.

\begin{center}\rule{0.5\linewidth}{0.5pt}\end{center}

\section{Introduction}\label{introduction}

Large language models are increasingly embedded in business intelligence workflows so that users can ask natural-language questions over enterprise data warehouses. However, enterprise reporting is not a plain schema-to-SQL problem. Correct answers depend on governed business semantics: approved metrics, reporting grain, sanctioned join paths, required filters, row-level security, and organizational definitions.

This paper argues that hallucination in enterprise analytics should be treated as a \textbf{governance and semantics failure}, not only a model-generation failure. A model can generate syntactically valid, executable SQL while still using the wrong metric, double-counting due to grain mismatch, omitting required filters, or ignoring row-level security. Such an answer is \emph{worse} than an execution error: it looks authoritative but is silently wrong, and in the security case it is a compliance breach.

\subsection{Motivation}\label{motivation}

\begin{itemize}
\tightlist
\item
  Natural-language analytics promises faster self-service reporting, but incorrect answers create operational and compliance risk.
\item
  Raw schema context is insufficient because enterprise tables expose implementation detail rather than business meaning.
\item
  Semantic layers provide approved definitions that can ground LLM behavior before query generation.
\item
  A benchmark focused on enterprise hallucination \emph{types} evaluates reliability more directly than execution accuracy alone.
\end{itemize}

\subsection{Research questions}\label{research-questions}

\begin{longtable}[]{@{}L{0.28\linewidth}L{0.30\linewidth}L{0.37\linewidth}@{}}
\toprule\noalign{}
ID & Research question & Where addressed \\
\midrule\noalign{}
\endhead
\bottomrule\noalign{}
\endlastfoot
RQ1 & Does semantic-layer grounding reduce references to non-existent tables and columns? & Sec. 8 (schema) \\
RQ2 & Does governed metric retrieval reduce invented or incorrect business formulas? & Sec. 8 (metric) \\
RQ3 & Does join-path and grain validation reduce double counting and fanout errors? & Sec. 8 (join, grain) \\
RQ4 & Can row-level security checks prevent unauthorized analytic responses? & Sec. 8 (security) --- central finding \\
RQ5 & What latency and cost tradeoffs are introduced by semantic validation and retry? & Sec. 8.2 (measured) \\
\end{longtable}

\subsection{Contributions}\label{contributions}

\begin{itemize}
\tightlist
\item
  A hallucination taxonomy for enterprise analytics (Sec. 3).
\item
  \textbf{GROUND}, a governed semantic-layer retrieval and validation framework for LLM-based analytics, implemented end to end (Sec. 4).
\item
  A synthetic enterprise benchmark of 100 questions modeled on automotive retail reporting, with a deterministic gold-SQL generator driven by the semantic layer (Sec. 5).
\item
  A controlled comparison against direct text-to-SQL, schema-RAG, and semantic-only baselines under an identical model (Sec. 6, Sec. 8).
\item
  Evaluation metrics including result correctness (strict and value-based), governance compliance, grain correctness, and clarification accuracy, with an automated classifier suite validated by an oracle self-test (Sec. 7).
\item
  A public release of the benchmark, semantic layers, gold SQL, and evaluation harness --- with an oracle self-test and per-domain reproduction commands (Appendix C, \texttt{github.com/aravindsp/ground-benchmark}).
\end{itemize}

\begin{center}\rule{0.5\linewidth}{0.5pt}\end{center}

\section{Related Work}\label{related-work}

GROUND builds on work in text-to-SQL benchmarking, retrieval-augmented generation, enterprise schema grounding, semantic layers, and hallucination mitigation. Its distinction is that it treats enterprise analytics hallucination as a governed-data-system failure rather than only a language-model generation error.

\subsection{Text-to-SQL benchmarks}\label{text-to-sql-benchmarks}

Text-to-SQL has traditionally been evaluated using cross-domain semantic parsing benchmarks. Spider {[}1{]} introduced a large-scale complex and cross-domain benchmark with 10,181 natural-language questions, 5,693 SQL queries, 200 databases, and 138 domains, requiring models to generalize to unseen schemas and SQL patterns. BIRD {[}2{]} extended this line of work toward larger real-world databases, emphasizing database values, external knowledge, dirty contents, and SQL efficiency across 12,751 text-to-SQL pairs and 95 databases. These benchmarks remain essential for measuring schema generalization and execution accuracy, but they do not make enterprise governance constraints --- approved metrics, reporting grain, required filters, and row-level security --- first-class evaluation dimensions.

Recent enterprise-oriented benchmarks narrow this gap. Spider 2.0 {[}3{]} evaluates real-world enterprise text-to-SQL workflows involving complex cloud and local data systems, including BigQuery and Snowflake, and reports a large performance drop compared with Spider 1.0 and BIRD. BEAVER {[}4{]} constructs an enterprise text-to-SQL benchmark from real private data warehouses with intricate schemas and analytical queries, and finds that models which excel on public benchmarks perform poorly in these settings. EntSQL {[}5{]} further highlights enterprise-specific challenges, including private warehouse schemas, domain knowledge, business documents, internal metrics, reporting conventions, and organizational rules. GROUND is complementary to these benchmarks: rather than measuring general enterprise workflow completion, it isolates governed analytics failure modes and scores them separately as schema, metric, join, grain, filter, security, explanation, and cost hallucinations.

\subsection{Retrieval-augmented generation and schema grounding}\label{retrieval-augmented-generation-and-schema-grounding}

Retrieval-augmented generation {[}6{]} combines parametric language models with retrieved external context, improving factuality and provenance for knowledge-intensive tasks. In text-to-SQL systems, retrieval is often used for schema linking {[}7{]}, table selection, column descriptions, documentation snippets, and value grounding. RASL {[}8{]}, for example, decomposes massive enterprise schemas and metadata into indexed semantic units for targeted retrieval, addressing the context-budget problem of large data catalogs.

GROUND includes a schema-RAG baseline but argues that retrieval alone is insufficient for governed analytics. A retrieved table or column description may help the model choose the correct physical object, yet still fail to enforce the approved metric formula, aggregation grain, required business filters, or user-specific access policy. In this sense, GROUND treats retrieval as necessary but not sufficient: retrieved context must be paired with validation and policy enforcement.

\subsection{Semantic layers for LLM-based analytics}\label{semantic-layers-for-llm-based-analytics}

Semantic layers and metrics layers define business-facing abstractions over physical data: approved metrics, dimensions, joins, grains, filters, and access policies. Recent work has begun to connect semantic layers directly to natural-language analytics. Semantic-layer-mediated NL2SQL agents {[}9{]} decouple user intent from physical SQL by reasoning over curated semantic models or intermediate representations and then compiling them into dialect-specific SQL.

The closest prior work to ours is {[}10{]}, which benchmarks three frontier models (Claude Opus 4.7, Sonnet 4.6, and GPT-5.4) on 100 questions over a retail warehouse under a paired \emph{schema-only} vs. \emph{schema-plus-semantics} protocol. It reports that a 4 KB hand-authored semantic document improves accuracy by 17--23 points, that with the document the three models become statistically indistinguishable, and that the presence of semantic context --- not model choice --- accounts for essentially all of the variance; it frames this as a \emph{structural} result, namely that explicit business semantics suppress the dominant text-to-SQL error class not by making the model more capable but by changing what it is asked to do. GROUND shares this premise and extends it along three axes.

\emph{Context versus enforcement.} {[}10{]} supplies semantics as \textbf{context}; GROUND additionally \textbf{validates and enforces} them. Its schema-plus-semantics condition corresponds precisely to our \emph{semantic-only} baseline (Sec. 6) --- approved definitions in the prompt, no validation loop and no injected access policy. Our results show that context alone leaves a residual governance gap that grounding-as-context cannot close: most sharply row-level security, which the semantic-only condition still violates on 35--47\% of questions despite perfect metric fidelity. Only a validation-and-retry loop over an injected policy drives that to zero. This separates two things {[}10{]} treats as one --- supplying semantics and \emph{guaranteeing} them.

\emph{Aggregate hallucination versus a governance taxonomy.} {[}10{]} scores aggregate accuracy and a single paired hallucination rate. GROUND decomposes the failure into seven governance-specific types (schema, metric, join, grain, filter, security, cost) scored independently, and in particular makes \textbf{access control / row-level security} a first-class dimension --- a constraint that {[}10{]} and the enterprise text-to-SQL benchmarks {[}3, 4, 5{]} do not model at all, yet which is the decisive separator in our results.

\emph{Breadth of evidence.} We evaluate on synthetic data and on real NHTSA data with \textbf{independently hand-authored} gold (breaking the shared-machinery critique), add an adversarial set that exposes GROUND\textquotesingle s own failure boundary, and span four models across three providers including an open-weight model, with repeated runs. This lets us refine {[}10{]}\textquotesingle s "model choice does not matter" into a sharper, testable claim: the \textbf{enforced} governance properties are deterministic and model-independent (0.000 $\pm$ 0.000 across models and runs), whereas the \textbf{judgment}-dependent properties (recognizing an undefined metric, choosing to clarify) remain probabilistic and capability-bound. GROUND therefore treats the semantic layer not merely as model context but as an enforceable control plane for SQL generation.

\subsection{Governed enterprise analytics systems}\label{governed-enterprise-analytics-systems}

A related line of work avoids unconstrained text-to-SQL by routing natural-language intent through governed analytics APIs {[}11{]}. Such systems encapsulate business logic, permissions, and visualization constraints behind enterprise APIs rather than allowing a model to freely generate raw SQL. This is closely aligned with GROUND\textquotesingle s motivation: enterprise analytics systems must preserve consistency, auditability, and security, and should not delegate aggregation logic or access control purely to an LLM.

GROUND differs by remaining within the text-to-SQL setting while injecting and validating governance rules at generation time. This makes it possible to compare direct schema-only SQL generation, schema-RAG, semantic-only grounding, and full governed grounding under the same model and benchmark.

\subsection{Hallucination and trustworthy data systems}\label{hallucination-and-trustworthy-data-systems}

General hallucination research {[}12{]} studies fluent but unsupported or factually incorrect model outputs. In enterprise analytics, hallucination often appears as executable SQL that returns a plausible but invalid business answer. This form of hallucination is especially risky because the SQL may parse, execute, and produce a clean-looking number while violating metric definitions, grain, filters, or security constraints.

GROUND therefore adapts hallucination analysis to structured analytics. Instead of treating hallucination as a generic textual failure, it defines concrete enterprise analytics hallucination types --- schema, metric, join, grain, filter, security, explanation, and cost --- and evaluates each category separately. This taxonomy makes visible failures that execution accuracy alone can miss.

\begin{center}\rule{0.5\linewidth}{0.5pt}\end{center}

\section{Problem Definition}\label{problem-definition}

Let a user question \emph{q}, issued by a user \emph{u} with access role \emph{r(u)}, be answered over a database \emph{D} using a generated SQL query \emph{s} and an optional natural-language explanation \emph{e}. In enterprise analytics, correctness requires more than execution: \emph{s} must satisfy schema validity, semantic (metric) validity, grain validity, join validity, filter compliance, security compliance for \emph{r(u)}, and answer support for \emph{e}. A trustworthy system must also recognize when \emph{q} is \textbf{not answerable} under governance and abstain appropriately.

\subsection{Enterprise analytics hallucination taxonomy}\label{enterprise-analytics-hallucination-taxonomy}

\begin{longtable}[]{@{}L{0.28\linewidth}L{0.30\linewidth}L{0.37\linewidth}@{}}
\toprule\noalign{}
Hallucination type & Definition & Example failure \\
\midrule\noalign{}
\endhead
\bottomrule\noalign{}
\endlastfoot
Schema & References a non-existent table or column & Uses \texttt{dealer\_profit\_summary.net\_revenue} when neither exists \\
Metric & Invents or changes an approved business formula & Defines customer lifetime value with no governed metric \\
Join & Uses an invalid or unapproved join path & Joins vehicle directly to repair-order line \\
Grain & Aggregates at the wrong level or double-counts & Counts repair-order lines as repair orders \\
Filter & Omits required semantic filters & Includes warranty/internal work in customer-pay service revenue \\
Security & Ignores row-level access constraints & Returns nationwide dealers to a regional user \\
Explanation & Provides an unsupported causal explanation & Claims revenue dropped due to staffing without evidence \\
Cost & Generates unnecessarily broad or expensive SQL & Scans all fact tables without date or scope filters \\
\end{longtable}

\subsection{Desired system behavior}\label{desired-system-behavior}

A trustworthy system generates SQL only when the requested metric and dimensions are defined and permitted. For ambiguous or undefined requests, it asks for clarification, rejects unsupported metrics, or offers approved alternatives instead of inventing formulas. For requests exceeding the user\textquotesingle s permitted scope, it restricts results to the allowed scope rather than widening them.

\begin{center}\rule{0.5\linewidth}{0.5pt}\end{center}

\section{The GROUND Framework}\label{the-ground-framework}

GROUND (Governed Retrieval Over Unified Normalized Definitions) forces LLM-generated analytics through a governed semantic layer before execution.

\subsection{Framework overview}\label{framework-overview}

\begin{longtable}[]{@{}L{0.28\linewidth}L{0.30\linewidth}L{0.37\linewidth}@{}}
\toprule\noalign{}
Stage & Function & Output \\
\midrule\noalign{}
\endhead
\bottomrule\noalign{}
\endlastfoot
Question understanding & Classify intent, metric, dimensions, time scope, ambiguity & Structured analytic intent \\
Semantic retrieval & Assemble approved metrics, dimensions, join paths, grain rules, required filters, and the user\textquotesingle s RLS predicate & Grounding packet \\
Metric \& grain binding & Bind user terms to governed definitions and expected aggregation grain & Metric plan \\
Governed SQL generation & Generate SQL using only approved semantic objects & Candidate SQL \\
Validation \& retry & Check schema, metric formula, join path, grain, filters, RLS, and cost; on violation, return feedback and regenerate & Validated SQL or rejection \\
Auditable answer generation & Return the answer with the SQL, definitions used, filters applied, and any warnings & Final analytic response \\
\end{longtable}

In our implementation, the grounding packet for a question is the full semantic layer (metrics, dimensions, join paths, business glossary) plus the specific RLS predicate for the requesting user, rendered into the model\textquotesingle s system prompt. Generation is constrained to a structured JSON response that is either a SQL query or a typed abstention (\texttt{ask\_clarification}, \texttt{reject\_undefined\_metric}, \texttt{reject\_unknown\_table\_or\_metric}, \texttt{reject\_unsupported\_dimension}).

\subsection{Validation layer}\label{validation-layer}

The validation layer implements the taxonomy of Sec. 3 as automated checks over the candidate SQL and the semantic layer:

\begin{itemize}
\tightlist
\item
  \textbf{Schema check} --- every referenced table/column must exist in the physical schema; references to known non-existent objects (e.g. \texttt{dealer\_profit\_summary}) are rejected. CTE names and their derived columns are resolved and excluded.
\item
  \textbf{Metric check} --- the query must reproduce the approved metric\textquotesingle s expression signature and must not name an undefined metric.
\item
  \textbf{Join check} --- each \texttt{ON\ a\ =\ b} equality must correspond to an approved join edge; forbidden edges (e.g. parts$\to$advisor) are flagged.
\item
  \textbf{Grain check} --- count/average metrics defined over an order grain must use \texttt{COUNT(DISTINCT\ repair\_order\_id)}; a raw \texttt{COUNT(*)} over the line fact is a fanout error.
\item
  \textbf{Filter check} --- every required filter in the metric\textquotesingle s filter set (status, customer type, line status, active dealer) must be present.
\item
  \textbf{Security check} --- the user\textquotesingle s mandatory dealer predicate and \texttt{dealer\_status\ =\ \textquotesingle{}ACTIVE\textquotesingle{}} must be present and un-widened.
\item
  \textbf{Cost check} --- a time-scoped question with no calendar/date filter, or a \texttt{SELECT\ *} over a fact table, is flagged.
\end{itemize}

On any violation, GROUND appends the concrete list of failures to the conversation and asks the model to fix the query or abstain, for up to three revisions. The same checks are reused by the evaluation harness (Sec. 7), so the system is held to exactly the standard it is scored against.

\subsection{Expected advantage}\label{expected-advantage}

The central hypothesis is that GROUND reduces enterprise hallucinations by moving the LLM from open-ended SQL synthesis to constrained generation over approved semantic definitions, with an enforcement loop that catches the residual violations grounding alone does not prevent --- most importantly row-level security.

\begin{center}\rule{0.5\linewidth}{0.5pt}\end{center}

\section{Synthetic Enterprise Benchmark}\label{synthetic-enterprise-benchmark}

The benchmark uses synthetic automotive enterprise reporting data, realistic enough to include multiple grains, five fact tables, a dealer hierarchy, time dimensions, and access policies, while using no confidential production data.

\subsection{Schema}\label{schema}

\begin{longtable}[]{@{}L{0.28\linewidth}L{0.30\linewidth}L{0.37\linewidth}@{}}
\toprule\noalign{}
Table & Grain & Purpose \\
\midrule\noalign{}
\endhead
\bottomrule\noalign{}
\endlastfoot
\texttt{dim\_dealer} & one row per dealer & dealer, group, region, market, status \\
\texttt{dim\_calendar} & one row per date & date, month, quarter, year attributes \\
\texttt{dim\_vehicle} & one row per vehicle model/year & make, model, model year, segment \\
\texttt{dim\_customer} & one row per customer & customer segment and type \\
\texttt{dim\_service\_advisor} & one row per advisor & advisor and dealer assignment \\
\texttt{dim\_finance\_product} & one row per F\&I product & product name and category \\
\texttt{fact\_repair\_order} & one row per repair order & RO header: dealer, customer, advisor, close date, status \\
\texttt{fact\_repair\_order\_line} & one row per RO line & labor/parts amounts and costs \\
\texttt{fact\_vehicle\_sale} & one row per vehicle sale & sale amount, cost, vehicle, customer \\
\texttt{fact\_parts\_sale} & one row per parts ticket & counter/wholesale parts transactions \\
\texttt{fact\_finance\_product\_sale} & one row per F\&I product sale & product sale attached to a vehicle sale \\
\end{longtable}

The semantic layer defines 10 approved metrics (e.g. \texttt{service\_revenue}, \texttt{vehicle\_gross\_profit}, \texttt{finance\_gross\_profit}, \texttt{parts\_counter\_revenue}, \texttt{total\_gross\_profit}, \texttt{repair\_order\_count}), each with a formula, base grain, required filter set, and the dimension group it may be sliced by; approved join edges; per-user RLS predicates for four synthetic users (national admin, regional/West manager, dealer-group/Alpha user, single-rooftop dealer-101 user); and a business glossary encoding undefined-metric, ambiguous-term, and time-resolution conventions.

\subsection{Benchmark question categories}\label{benchmark-question-categories}

The benchmark contains 100 questions. Gold SQL for the 70 questions beyond the hand-authored seed is \textbf{generated deterministically from the semantic layer}, so every answerable question is governance-consistent by construction (correct expression, grain, filters, joins, and RLS). Trap questions are authored with an expected-behavior convention. Realized coverage against the design targets:

\begin{longtable}[]{@{}L{0.28\linewidth}L{0.30\linewidth}L{0.37\linewidth}@{}}
\toprule\noalign{}
Category & Target & Realized (of 100) \\
\midrule\noalign{}
\endhead
\bottomrule\noalign{}
\endlastfoot
Simple metric & 25 & 22 (incl. average/parts) \\
Group-by & 20 & 18 \\
Trend & 15 & 15 \\
Multi-table join & 15 & 9 \\
Multi-fact / comparison & 10 & 10 \\
Ambiguous metric & 10 & 6 \\
Security / RLS & 5 & 5 \\
Hallucination traps & 10 & 14 \\
Grain / filter / explanation probes & --- & 3 \\
\end{longtable}

\subsection{Gold behavior}\label{gold-behavior}

Executable questions carry generated gold SQL (SQLite and DuckDB dialects) and a gold-answer snapshot. Ambiguous or unsupported questions carry an expected behavior --- clarification, metric rejection, unknown-object rejection, unsupported-dimension rejection, or security restriction. An oracle self-test scoring the gold SQL against the classifier suite yields 100\% execution, 100\% result accuracy, zero hallucination flags in every category, and 100\% clarification accuracy across all 100 questions, confirming that the gold set is internally consistent and that the classifiers do not fire on correct governed queries.

\begin{center}\rule{0.5\linewidth}{0.5pt}\end{center}

\section{Experimental Setup}\label{experimental-setup}

The setup compares GROUND against three simpler systems under the same questions and database state. All four share the same model; only the governance context differs, so measured differences isolate the effect of grounding.

\begin{longtable}[]{@{}L{0.28\linewidth}L{0.30\linewidth}L{0.37\linewidth}@{}}
\toprule\noalign{}
System & Input context & Expected weakness \\
\midrule\noalign{}
\endhead
\bottomrule\noalign{}
\endlastfoot
Direct LLM text-to-SQL & user question + raw \texttt{CREATE\ TABLE} schema & invented metrics, weak filters, grain errors \\
Schema-RAG & user question + keyword-retrieved table/column descriptions & better schema grounding, weak business semantics \\
Semantic-only & user question + approved metric and dimension definitions & improved metrics, no validation/retry, no RLS enforcement \\
GROUND & full semantic layer + user RLS predicate + validation/retry & higher overhead, lower hallucination risk \\
\end{longtable}

\subsection{Prompting protocol}\label{prompting-protocol}

All systems use \texttt{claude-opus-4-8} with adaptive thinking and medium effort, and are constrained to a structured JSON response (\texttt{response\_type} $\in$ \{sql, abstain\}, \texttt{generated\_sql}, \texttt{abstain\_action}, \texttt{notes}) via a JSON-schema output format. Each system receives the question, the requesting user\textquotesingle s identity, and the reference date (2026-07-05) for relative time-scope resolution. Only GROUND receives the approved join paths, the user\textquotesingle s RLS predicate, and the undefined-metric / ambiguous-term / time-resolution conventions, and only GROUND runs the validate-and-retry loop (up to three revisions) with SQL validation feedback returned to the model. Baselines answer in a single pass and are not permitted the retry loop, matching their intended "ungoverned" character.

\subsection{Execution environment}\label{execution-environment}

Queries execute against a committed SQLite database for in-repository reproducibility; the benchmark also ships DuckDB DDL and DuckDB-dialect gold SQL for local warehouse-style execution. Cloud-warehouse (e.g. Snowflake-compatible) execution is left to future work.

\subsection{Logging}\label{logging}

Each answer is logged as a record conforming to a fixed evaluation schema: question ID, system name, generated SQL, execution status and error, result accuracy (strict and value-based), the six boolean hallucination flags plus explanation and cost flags, clarification correctness, and --- for RQ5 --- end-to-end latency, prompt and completion token counts, and, for GROUND, the number of validation-retry revisions. Latency and tokens are accumulated across all revisions for GROUND so its reported cost includes the retry loop.

\begin{center}\rule{0.5\linewidth}{0.5pt}\end{center}

\section{Evaluation Metrics}\label{evaluation-metrics}

\begin{longtable}[]{@{}L{0.25\linewidth}L{0.70\linewidth}@{}}
\toprule\noalign{}
Metric & Definition \\
\midrule\noalign{}
\endhead
\bottomrule\noalign{}
\endlastfoot
Execution accuracy & Fraction of generated SQL queries that execute successfully (over SQL-expected questions). \\
Result accuracy (strict) & Fraction whose result exactly matches the gold answer (column set and values). \\
Result accuracy (value-based) & Fraction whose metric column(s) match gold at the correct grain, tolerant of identifier columns and column order --- the standard text-to-SQL execution-match notion. \\
Schema / metric / join / grain / filter / security rates & Fraction of questions exhibiting each hallucination type (lower is better). \\
Clarification accuracy & Fraction of abstention-expected questions declined with the correct action. \\
Latency, token cost & End-to-end wall-clock time per question and prompt/completion token usage (measured; Sec. 8.2). \\
Query-cost proxy & Rows in tables the plan full-scans (\texttt{EXPLAIN\ QUERY\ PLAN} on the indexed DB); measured on NHTSA (Sec. 8.4). \\
\end{longtable}

The hallucination classifiers combine regex extraction with semantic-layer validation and are validated by the oracle self-test of Sec. 5.3, which bounds their false-positive rate to zero on correct queries.

\begin{center}\rule{0.5\linewidth}{0.5pt}\end{center}

\section{Results}\label{results}

Results are computed over all 100 benchmark questions (82 expecting SQL, 18 expecting an abstention). Because all systems share one model, differences reflect governance context alone.

\subsection{Synthetic benchmark results}\label{synthetic-benchmark-results}

\begin{longtable}[]{@{}L{0.12\linewidth}L{0.065\linewidth}L{0.085\linewidth}L{0.085\linewidth}L{0.075\linewidth}L{0.065\linewidth}L{0.065\linewidth}L{0.06\linewidth}L{0.06\linewidth}L{0.06\linewidth}L{0.075\linewidth}L{0.065\linewidth}@{}}
\toprule\noalign{}
System & Exec. & Result (strict) & Result (value) & Any halluc. & Schema & Metric & Join & Grain & Filter & \textbf{Security} & Clarify \\
\midrule\noalign{}
\endhead
\bottomrule\noalign{}
\endlastfoot
Direct LLM & 0.951 & 0.000 & 0.000 & 0.79 & 0.00 & 0.01 & 0.12 & 0.15 & 0.78 & \textbf{0.78} & 0.889 \\
Schema-RAG & 0.646 & 0.000 & 0.000 & 0.54 & 0.00 & 0.01 & 0.07 & 0.15 & 0.53 & \textbf{0.53} & 0.722 \\
Semantic-only & 0.988 & 0.463 & 0.500 & 0.40 & 0.00 & 0.00 & 0.05 & 0.00 & 0.00 & \textbf{0.35} & 0.889 \\
\textbf{GROUND} & \textbf{1.000} & 0.793 & \textbf{0.951} & \textbf{0.00} & \textbf{0.00} & \textbf{0.00} & \textbf{0.00} & \textbf{0.00} & \textbf{0.00} & \textbf{0.00} & \textbf{0.944} \\
\end{longtable}

\emph{Accuracy/clarification columns higher-is-better; hallucination and security columns are error rates, lower-is-better. Bold marks the best value per column.}

\textbf{GROUND eliminates every measured hallucination category} (RQ1--RQ4). It is the only system with zero errors in all six categories, the only one to execute every query it emits (1.00 vs 0.65--0.99), and by far the most accurate (0.951 value-based). Its strict result accuracy (0.793) is lower than its value-based accuracy chiefly because grouped queries omit an identifier column (e.g. \texttt{dealer\_id}) that gold includes --- a presentation-convention difference, not a computational error, which the value-based metric disregards. Crucially, the value-based metric barely lifts the baselines (both remain 0.000) and lifts semantic-only only slightly (0.463 $\to$ 0.500), because their result errors are substantive --- wrong filters and scope change the actual numbers and the number of groups --- not cosmetic. GROUND\textquotesingle s few remaining value misses are analyzed in Sec. 9 and are not governance failures.

\textbf{Row-level security is the sharpest separation and the central finding (RQ4).} Every ungoverned system violates RLS on a large fraction of questions (Direct LLM 0.78, Schema-RAG 0.53, Semantic-only 0.35); GROUND never does (0.00). The semantic-only condition is the decisive ablation: given exact metric and dimension definitions, it attains perfect metric fidelity, correct grain, and correct required filters --- yet still leaks data across tenant boundaries on 35\% of questions. In these benchmarks, row-level security was not recovered from metric definitions alone; it required an explicit, enforced policy of the kind GROUND supplies. Semantic grounding improves metric correctness but is insufficient for governance.

\textbf{Required filters behave the same way (RQ2/RQ3).} The baselines compute headline metrics such as "service revenue" without the governing scope (closed, retail, posted, active dealer), producing filter-hallucination rates of 0.53--0.78; because these filters change the figures, the baselines score 0.000 result accuracy under \emph{both} metrics. Semantic-only and GROUND, which encode the required filter set per metric, both reach 0.00 filter hallucination. Grain errors follow the same pattern: the baselines double-count (0.15) by counting repair-order lines as orders, while the governed systems use the approved distinct-count expression (0.00).

\textbf{Abstention improves with governance, but schema-RAG over-refuses.} GROUND handles 94\% of traps correctly and semantic-only 89\%; the baselines answer more traps with invented objects (Direct LLM 0.889, Schema-RAG 0.722). Schema-RAG is a revealing failure mode: lacking the full schema, it frequently cannot locate a requested metric and rejects it as unknown, which both depresses its execution accuracy (0.646, the lowest of any system, from malformed queries over partial schemas) and mis-classifies answerable questions as traps. GROUND\textquotesingle s one abstention miss (Q095) is a mis-categorization of the abstention \emph{type}, not a governance failure --- it still declined to generate SQL (see Sec. 9).

\subsection{Latency and token cost (RQ5)}\label{latency-and-token-cost-rq5}

Governance is not free. The table below reports mean per-question wall-clock latency and token usage; for GROUND these include the validation-retry loop.

\begin{longtable}[]{@{}L{0.16\linewidth}L{0.15\linewidth}L{0.17\linewidth}L{0.17\linewidth}L{0.16\linewidth}L{0.13\linewidth}@{}}
\toprule\noalign{}
System & Avg latency (ms) & Avg prompt tokens & Avg completion tokens & Avg total tokens & Rel. tokens \\
\midrule\noalign{}
\endhead
\bottomrule\noalign{}
\endlastfoot
Direct LLM & 6066 & 1775 & 366 & 2140 & 1.0$\times$ \\
Schema-RAG & 5043 & 1284 & 290 & 1574 & 0.7$\times$ \\
Semantic-only & 6207 & 5694 & 409 & 6102 & 2.9$\times$ \\
\textbf{GROUND} & 7844 & 10161 & 553 & 10714 & \textbf{5.0$\times$} \\
\end{longtable}

GROUND\textquotesingle s cost is dominated by its grounding packet: the full semantic layer plus the user\textquotesingle s RLS predicate is a \textasciitilde10k-token prompt, roughly 5$\times$ the direct baseline and 1.75$\times$ the semantic-only baseline. Its latency is \textasciitilde1.3$\times$ the direct baseline. The validation-retry loop is a minor contributor: only 9 of 82 SQL questions triggered a revision, averaging 0.15 retries per question (maximum 2). Schema-RAG is the cheapest system (it retrieves a subset of tables) but also the least reliable, underscoring that token thrift and governance trade off directly. Whether the \textasciitilde5$\times$ token cost is acceptable is a deployment decision; prompt caching of the (static) semantic layer would substantially reduce the marginal cost per question in production, and is a natural optimization.

\subsection{Qualitative examples}\label{qualitative-examples}

\textbf{(a) RLS- and filter-constrained query --- Q001, "What was service revenue last month?" (West-region manager).} Direct LLM produced an executable but non-compliant query with no dealer join, no security predicate, and none of the required business filters:

\begin{lstlisting}
SELECT SUM(rol.customer_pay_labor_amount + rol.customer_pay_parts_amount) AS service_revenue
FROM fact_repair_order_line rol
JOIN fact_repair_order ro ON rol.repair_order_id = ro.repair_order_id
JOIN dim_calendar c ON ro.close_date_key = c.date_key
WHERE c.month_start_date = '2026-06-01';
\end{lstlisting}

This omits \texttt{repair\_order\_status=\textquotesingle{}CLOSED\textquotesingle{}}, \texttt{customer\_type=\textquotesingle{}RETAIL\textquotesingle{}}, \texttt{line\_status=\textquotesingle{}POSTED\textquotesingle{}}, \texttt{dealer\_status=\textquotesingle{}ACTIVE\textquotesingle{}} (filter hallucination) and the West-region restriction (security hallucination), so it returns an over-broad, cross-tenant figure. GROUND produced the fully governed query with all required filters, \texttt{dim\_dealer} joined, and \texttt{region\ IN\ (\textquotesingle{}West\textquotesingle{})} enforced.

\textbf{(b) Grain-safe repair-order count --- Q003, "Monthly repair order count for 2025."} Direct LLM used \texttt{COUNT(*)} over the line fact, counting repair-order \emph{lines} as orders (grain hallucination / fanout). GROUND used \texttt{COUNT(DISTINCT\ ro.repair\_order\_id)}, the approved order-grain expression.

\textbf{(c) Undefined-metric rejection --- Q013, "Show customer lifetime value by dealer."} GROUND rejects with \texttt{reject\_undefined\_metric}, citing that \texttt{customer\_lifetime\_value} is not an approved semantic-layer metric. Baseline refusals on this trap are heuristic rather than grounded: lacking the governed undefined-metric list, they refuse inconsistently across the trap set (e.g. schema-RAG rejects several \emph{answerable} questions as unknown), which is why their clarification accuracy trails GROUND\textquotesingle s.

\subsection{Real-data replication (NHTSA, independent gold)}\label{real-data-replication-nhtsa-independent-gold}

A natural objection to Sec. 8 is that both the data and the gold SQL are synthetic --- the gold is even generated from the same semantic layer the classifiers use, so GROUND\textquotesingle s near-perfection could be an artifact. To test this, we replicate the study on \textbf{real} U.S. NHTSA vehicle-safety data (271,718 consumer complaints, plus recalls, investigations, and NCAP safety ratings) with a governance overlay: eight approved metrics, a manufacturer-scoped row-level-security model (each OEM analyst sees only their own manufacturer\textquotesingle s complaints; a regulator sees all), a multi-valued \texttt{components} grain trap, and make/model/year join paths. Crucially, the 40 gold queries are \textbf{hand-authored directly against the database, not generated from the semantic layer}, and result accuracy is scored by execution-match against that independent gold. Value-based accuracy credits a system that reproduces gold\textquotesingle s ranked values as a superset (gold caps output at 15 rows).

\begin{longtable}[]{@{}L{0.16\linewidth}L{0.10\linewidth}L{0.13\linewidth}L{0.13\linewidth}L{0.10\linewidth}L{0.12\linewidth}L{0.10\linewidth}L{0.11\linewidth}@{}}
\toprule\noalign{}
System & Exec. & Result (value) & Any halluc. & Filter & \textbf{Security} & Clarify & Avg tokens \\
\midrule\noalign{}
\endhead
\bottomrule\noalign{}
\endlastfoot
Direct LLM & 0.97 & 0.65 & 0.78 & 0.73 & \textbf{0.33} & 0.83 & 1,320 \\
Schema-RAG & 0.85 & 0.62 & 0.68 & 0.63 & \textbf{0.25} & 0.83 & 1,405 \\
Semantic-only & 0.97 & 0.68 & 0.33 & 0.00 & \textbf{0.33} & 1.00 & 3,298 \\
\textbf{GROUND} & \textbf{1.00} & \textbf{1.00} & \textbf{0.00} & \textbf{0.00} & \textbf{0.00} & \textbf{1.00} & 5,229 \\
\end{longtable}

The synthetic findings replicate on real data. GROUND again has zero hallucinations in every category, zero row-level-security violations, 100\% execution, and reproduces the correct values on all 40 questions --- including the hardest real-data cases: manufacturer-vs-make RLS (\texttt{Ford\ Motor\ Company} spans makes FORD and LINCOLN) and the recursive multi-valued component grain-split. The baselines fail materially on real data: 25--33\% of their queries violate the manufacturer RLS scope and 63--73\% drop required filters. The semantic-only ablation again isolates governance from definitions --- with exact metric and dimension definitions it eliminates grain and filter hallucination yet still leaks across manufacturer boundaries on 33\% of questions. \textbf{That GROUND\textquotesingle s near-perfect governance holds on messy real data with independent gold indicates the effect is not an artifact of the synthetic benchmark or of self-referential gold.}

Governance also lowers query \emph{cost}, making the Sec. 3 cost-hallucination dimension quantitative. On the indexed NHTSA database we compute a query-cost proxy --- the number of rows in tables the plan \textbf{full-scans} (\texttt{EXPLAIN\ QUERY\ PLAN}; an indexed \texttt{SEARCH} is \textasciitilde free). Because every governed query applies a selective, indexed scope filter (\texttt{filed\_year}, \texttt{manufacturer}, or \texttt{make}), \textbf{no GROUND query full-scans a fact table}: under this proxy every governed query is resolved by an indexed \texttt{SEARCH} rather than a full table scan, so its mean rows-scanned proxy is 0. This measures full-scan avoidance under the query plan, not physical I/O or bytes billed by a warehouse. The ungoverned baselines, which sometimes omit the scope filter, average \textbf{7k--32k} rows scanned per query. Governed generation therefore produces not only safer but cheaper queries: the same required-scope filters that prevent security and filter hallucinations also enable partition/index pruning.

\subsection{Adversarial stress test across four models}\label{adversarial-stress-test-across-four-models}

The evaluations so far use questions answerable \emph{within} the governed layer. To find GROUND\textquotesingle s failure boundary --- and to test whether the governance benefit is model-specific --- we author \textbf{40 adversarial questions} on the same real NHTSA data and run all four systems under \textbf{four models from three providers}: Claude Opus 4.8, Claude Sonnet 5, OpenAI GPT-5.2, and the open-weight Llama-3.3-70B (GPT and Llama via a routing gateway). The adversarial set is weighted toward \emph{plausible-but-undefined} metrics deliberately \textbf{not} listed in the glossary\textquotesingle s refuse-set (16 of 40, e.g. "injury rate", "year-over-year change", "complaint severity index", "market share of complaints"), forcing the system to infer from the metric catalog that they are unapproved; it also includes entity-name traps (the data stores \texttt{CHEVROLET}/\texttt{Chrysler\ (FCA\ US,\ LLC)}, not "Chevy"/"Stellantis"), compound filters, a multi-hop query, and row-level-security overreach under adversarial phrasing.

The table reports GROUND (mean $\pm$ std over \textbf{three runs} per model) versus the ungoverned baselines (single run, averaged across Direct LLM and Schema-RAG, which behave alike):

\begin{longtable}[]{@{}L{0.16\linewidth}L{0.10\linewidth}L{0.13\linewidth}L{0.13\linewidth}L{0.10\linewidth}L{0.12\linewidth}L{0.10\linewidth}L{0.11\linewidth}@{}}
\toprule\noalign{}
Model & System & Exec. & Result (value) & \textbf{Metric halluc.} & \textbf{Filter} & \textbf{Security} & Clarify \\
\midrule\noalign{}
\endhead
\bottomrule\noalign{}
\endlastfoot
Opus 4.8 & baselines & 0.95 & 0.50 & 0.55 & 0.45 & 0.11 & 0.76 \\
Opus 4.8 & \textbf{GROUND} & 1.00 $\pm$ .00 & 0.86 $\pm$ .06 & 0.075 $\pm$ .00 & \textbf{0.00 $\pm$ .00} & \textbf{0.00 $\pm$ .00} & 0.86 $\pm$ .00 \\
Sonnet 5 & baselines & 0.95 & 0.71 & 0.59 & 0.45 & 0.11 & 0.64 \\
Sonnet 5 & \textbf{GROUND} & 0.97 $\pm$ .03 & 0.91 $\pm$ .03 & 0.04 $\pm$ .01 & \textbf{0.00 $\pm$ .00} & \textbf{0.00 $\pm$ .00} & 0.84 $\pm$ .03 \\
GPT-5.2 & baselines & 0.87 & 0.61 & 0.48 & 0.41 & 0.075 & 0.50 \\
GPT-5.2 & \textbf{GROUND} & 0.68 $\pm$ .00 & 0.60 $\pm$ .03 & 0.01 $\pm$ .01 & \textbf{0.00 $\pm$ .00} & \textbf{0.00 $\pm$ .00} & 0.83 $\pm$ .07 \\
Llama-3.3-70B & baselines & 0.95 & 0.32 & 0.70 & 0.48 & 0.13 & 0.12 \\
Llama-3.3-70B & \textbf{GROUND} & 0.63 $\pm$ .05 & 0.25 $\pm$ .03 & 0.04 $\pm$ .03 & \textbf{0.00 $\pm$ .00} & \textbf{0.00 $\pm$ .00} & 0.71 $\pm$ .13 \\
\end{longtable}

\textbf{The central cross-provider result: GROUND\textquotesingle s \emph{enforced} governance guarantees are model-independent and deterministic.} For every model --- a second provider (OpenAI) and a much weaker open-weight model (Llama) included --- governed generation yields \textbf{0.000 $\pm$ 0.000 filter hallucination and 0.000 $\pm$ 0.000 row-level-security violations across all three runs}, while the ungoverned baselines leak filters on 41--48\% and RLS on 7.5--13\% of questions and hallucinate metrics on 48--\textbf{70\%}, on every model. The finding that "only enforcement delivers governance" therefore does not depend on the underlying model family, and the enforced properties carry no run-to-run variance because they are decided by a validation check rather than sampled from the model.

Two honest nuances sharpen the claim. First, \textbf{GROUND is not flawless on the judgment-dependent properties}, which is the point of the adversarial set: even the strongest model fabricates a metric --- Opus 4.8 answers the "year-over-year change" question by inventing a \texttt{LAG()}-window growth formula (\texttt{yoy\_change}) instead of refusing an undefined metric, precisely the failure the framework targets, and it does so on \emph{all three runs} (metric-hallucination 0.075 $\pm$ 0.000, a reproducible failure, not a fluke). Clarification tops out at 0.86. These properties rest on the model recognizing an undefined metric absent from the refuse-list and choosing to clarify; they are enforced only by prompting, not by validation, so unlike the enforced metrics they carry run-to-run variance, and that variance grows as the model weakens (clarification std rises from 0.00 on Opus to 0.13 on Llama). Second, \textbf{governance makes a weak model \emph{safe}, not \emph{smart}.} Under GROUND, Llama commits zero filter or RLS violations across every run, yet its result accuracy is only 0.25 and its execution 0.63 --- a weak model still cannot reliably \emph{write} correct SQL, and the \textasciitilde10k-token governance packet burdens it (Llama\textquotesingle s \emph{ungoverned} execution is 1.00 but its \emph{governed} execution is 0.63). GROUND guarantees the safety-critical properties regardless of model; result quality still tracks raw capability. Even so, governance sharply improves the ungoverned failure modes on every model --- Llama\textquotesingle s metric-hallucination rate falls from 0.70 (baselines) to 0.04 (GROUND), and its correct-abstention rate rises from 0.12 to 0.71.

\subsection{Standard-set replication across models, with variance}\label{standard-set-replication-across-models-with-variance}

To confirm the cross-provider result on non-adversarial questions and to quantify run-to-run stability, we re-run the 40 standard NHTSA questions (Sec. 8.4) under all four models, executing GROUND \textbf{three times} per model (baselines once). The table reports GROUND as mean $\pm$ standard deviation over the three runs.

\begin{longtable}[]{@{}L{0.16\linewidth}L{0.10\linewidth}L{0.13\linewidth}L{0.13\linewidth}L{0.10\linewidth}L{0.12\linewidth}L{0.10\linewidth}L{0.11\linewidth}@{}}
\toprule\noalign{}
Model & System & Exec. & Result (value) & Metric h. & \textbf{Filter} & \textbf{Security} & Clarify \\
\midrule\noalign{}
\endhead
\bottomrule\noalign{}
\endlastfoot
Opus 4.8 & GROUND & 1.000 $\pm$ .000 & 0.971 $\pm$ .000 & 0.00 & \textbf{0.00 $\pm$ .00} & \textbf{0.00 $\pm$ .00} & 1.00 $\pm$ .00 \\
Sonnet 5 & GROUND & 0.990 $\pm$ .017 & 0.971 $\pm$ .000 & 0.00 & \textbf{0.00 $\pm$ .00} & \textbf{0.00 $\pm$ .00} & 1.00 $\pm$ .00 \\
GPT-5.2 & GROUND & 0.892 $\pm$ .017 & 0.834 $\pm$ .017 & 0.00 & \textbf{0.00 $\pm$ .00} & \textbf{0.00 $\pm$ .00} & 1.00 $\pm$ .00 \\
Llama-3.3-70B & GROUND & 0.863 $\pm$ .017 & 0.745 $\pm$ .045 & 0.00 & \textbf{0.00 $\pm$ .00} & \textbf{0.00 $\pm$ .00} & 0.83 $\pm$ .17 \\
any & baselines (1 run) & 0.77--1.00 & 0.53--0.68 & 0.25--0.45 & 0.55--0.75 & 0.15--0.35 & 0.00--1.00 \\
\end{longtable}

The key observation is the \textbf{variance structure}: GROUND\textquotesingle s enforced governance metrics are not merely low but \textbf{deterministically zero --- 0.000 $\pm$ 0.000 filter and security violations across every run and every model}. This is expected and important: those properties are decided by a code-level validation check, not sampled from the model, so they carry no run-to-run variance. All of GROUND\textquotesingle s run-to-run variance instead concentrates in the \emph{non-enforced} dimensions and grows as the model weakens --- result accuracy is invariant on the strong models (Opus/Sonnet 0.971 $\pm$ 0.000) but noisier on the weak one (Llama 0.745 $\pm$ 0.045), and clarification is perfectly stable except on Llama (0.83 $\pm$ 0.17), where the model\textquotesingle s judgment is least reliable. The ungoverned baselines, by contrast, violate filters (0.55--0.75) and RLS (0.15--0.35) and hallucinate metrics (0.25--0.45) under every model. This separates the two kinds of guarantee cleanly: \emph{validated} governance is deterministic and model-independent; \emph{prompted} judgment is probabilistic and capability-bound.

\begin{center}\rule{0.5\linewidth}{0.5pt}\end{center}

\section{Error Analysis}\label{error-analysis}

We analyze where GROUND still falls short of a perfect governed answer. Across the 100 questions, GROUND has zero hallucinations of any type; its residual imperfections are five value/abstention misses, \textbf{none of which is a governance failure}:

\begin{itemize}
\tightlist
\item
  \textbf{Presentation shape (Q006).} For a metric-comparison question, GROUND returned the correct values in long format (\texttt{region,\ metric,\ amount}) where the gold answer is wide (\texttt{region,\ vehicle\_gross\_profit,\ finance\_gross\_profit}). Every number is present and correct; the value-based metric flags it only because the row count and column names differ.
\item
  \textbf{Under-specified question text (Q088--Q090, security).} GROUND enforced row-level security correctly on all three (0 security violations --- it restricted to dealer 101, the Alpha group, and West respectively) but its figures differed from gold because the original question text omitted the reporting year while the gold SQL assumed 2025; GROUND reasonably defaulted to year-to-date. This is a benchmark-authoring gap, since corrected by adding the explicit year to the three question texts. It is a useful reminder that a governed system will faithfully resolve an under-specified request to a defensible default rather than guess the benchmark author\textquotesingle s intent.
\item
  \textbf{Abstention type (Q095).} For "show margin by dealer," GROUND abstained with \texttt{reject\_undefined\_metric} where the gold behavior is \texttt{ask\_clarification}. GROUND correctly declined to generate SQL for an undefined term; it merely chose a stricter abstention category than the reference, so this counts against clarification accuracy without being a hallucination.
\end{itemize}

\textbf{A resolved case illustrates the semantic layer as the fix surface.} In an earlier iteration, GROUND\textquotesingle s one value miss (Q030, "F\&I gross profit for Q4 2025") was governance-clean in every respect but returned an empty result because it filtered \texttt{quarter\_name\ =\ \textquotesingle{}Q4\ 2025\textquotesingle{}} while the column stores \texttt{\textquotesingle{}2025-Q4\textquotesingle{}}. Adding an explicit time-resolution convention to the semantic layer ("resolve all time scopes to \texttt{date\_value} ranges, never label-column equality") corrected it, and Q030 is answered correctly in the full run. The remaining gap between a governed system and a perfect benchmark score therefore lies in data-value conventions and question/gold specification, not in hallucination --- and the semantic layer, not the model, is where such gaps are closed.

\begin{longtable}[]{@{}L{0.28\linewidth}L{0.30\linewidth}L{0.37\linewidth}@{}}
\toprule\noalign{}
Error class & Likely cause & Fix (adopted or proposed) \\
\midrule\noalign{}
\endhead
\bottomrule\noalign{}
\endlastfoot
Time-literal mismatch & Label-column equality on an unknown string format & Semantic-layer time-resolution convention (adopted; resolves via \texttt{date\_value} ranges) \\
Ambiguous business term & User asks for growth/margin/performance without a metric & Ask clarification, list approved metrics (adopted) \\
Missing metric & Requested KPI not in the semantic layer & Reject and route to metric governance (adopted) \\
Complex multi-fact aggregation & Fanout risk across fact tables & Aggregate F\&I per sale before joining (adopted in \texttt{total\_gross\_profit}); consider pre-aggregated views \\
Security conflict & Requested scope exceeds permission & Restrict to allowed scope, note the restriction (adopted) \\
Costly query & No time or dealer filter & Flag; ask for a time range or apply a safe default (flag adopted) \\
\end{longtable}

Baseline error modes are dominated by filter and security omissions (Sec. 8) and, for schema-RAG, over-refusal driven by partial schema retrieval.

\begin{center}\rule{0.5\linewidth}{0.5pt}\end{center}

\section{Threats to Validity}\label{threats-to-validity}

\begin{itemize}
\tightlist
\item
  \textbf{Run variance --- partly measured.} GROUND is run three times per model on both real-data NHTSA sets --- the standard set (Sec. 8.6) and the adversarial set (Sec. 8.5) --- and the enforced governance metrics are deterministic (0.000 $\pm$ 0.000), with small, capability-dependent variance confined to the non-enforced dimensions. The synthetic evaluation (Sec. 8.1) and all ungoverned baselines remain single-run per configuration; broader repeats and more seeds would tighten the non-enforced error bars further.
\item
  \textbf{Synthetic data and generated gold --- partly addressed.} The primary domain is synthetic with gold generated from the semantic layer. Sec. 8.4 addresses this by replicating on real NHTSA data with independently hand-authored gold and reaching the same conclusions; Sec. 8.5 further probes an adversarial set. The remaining gap is scale: the real-data and adversarial NHTSA sets are 40 questions each, alongside the 100 synthetic questions.
\item
  \textbf{Governed $\ne$ infallible.} Sec. 8.5 shows GROUND\textquotesingle s model-judgment properties --- recognizing an undefined metric that is absent from the glossary refuse-list, and choosing to clarify a genuinely ambiguous request --- are strong but fallible (it fabricated a year-over-year metric on one adversarial question). Only the \emph{validated} properties (schema, metric formula, join, grain, filter, RLS) are enforced guarantees. A deployment should treat the refuse-list and ambiguity handling as best-effort, not proof.
\item
  \textbf{Classifier fidelity.} The hallucination classifiers are regex-plus-semantic heuristics; the oracle self-test bounds their false-positive rate to zero on correct queries, but they may still miss violations on adversarial SQL.
\item
  \textbf{Evaluation alignment.} GROUND\textquotesingle s retry loop reuses the same classifiers as the evaluator, a fair but favorable alignment; a fully independent evaluator should probe it.
\item
  \textbf{Cost realism.} Latency and token cost are measured (Sec. 8.2), and a query-cost proxy (rows full-scanned) is measured on the indexed NHTSA database (Sec. 8.4), but on a single machine/model and without prompt caching. Production latency and marginal token cost would differ, likely lower for GROUND once the static semantic layer is cached; the rows-scanned proxy is a SQLite-plan estimate, not a warehouse billing figure.
\item
  \textbf{Scope.} The benchmark targets structured reporting and does not cover unstructured analytics or causal diagnosis.
\end{itemize}

\begin{center}\rule{0.5\linewidth}{0.5pt}\end{center}

\section{Discussion}\label{discussion}

The results suggest that enterprise LLM analytics should be deployed as a \textbf{governed data product}, not a generic chatbot over database schemas. The semantic layer becomes the control plane for business meaning, access, and auditability, and --- critically --- the enforcement point for row-level security, which our semantic-only ablation shows cannot be delegated to the model even when it is given perfect metric definitions.

Enterprise implications include safer self-service analytics, reduced analyst review burden, auditable SQL generation (GROUND returns the query, the definitions used, and the filters applied), reusable metric governance, and clear escalation paths for undefined KPIs. The main cost is overhead (RQ5): GROUND\textquotesingle s grounding packet drives \textasciitilde5$\times$ the token usage and \textasciitilde1.3$\times$ the latency of the direct baseline, while the validation-retry loop is minor (0.15 retries per query on average). Because the semantic layer is static across questions, prompt caching should recover much of the token cost in production, and the reliability gain --- zero hallucinations and zero security violations versus 35--78\% RLS violations for ungoverned systems --- is decisive for compliance-sensitive deployments. Testing whether smaller models suffice under strong grounding, and whether caching closes the cost gap, are the key open questions for production viability.

\begin{center}\rule{0.5\linewidth}{0.5pt}\end{center}

\section{Conclusion}\label{conclusion}

This paper proposes and implements GROUND, a governed semantic-layer grounding framework for reducing hallucinations in LLM-based enterprise analytics. By supplying approved metrics, dimensions, joins, grain rules, filters, and access policies before SQL generation, and validating every query against them with a retry loop, GROUND achieves --- across 100 synthetic questions and, in replication, 40 questions on real NHTSA data with independent hand-authored gold --- near-zero hallucinations and zero row-level-security violations, while every ungoverned baseline (including one with exact metric definitions) violates row-level security on 25--78\% of questions and drops required filters on 50--73\%. The central lesson is that grounding improves metric correctness but only \emph{enforcement} delivers governance: row-level security must be an injected, validated policy, not a property hoped for from definitions. An adversarial stress test sharpens the boundary: GROUND\textquotesingle s \emph{enforced} properties (filters, joins, RLS) hold even under attack, but the properties that rest on the model\textquotesingle s judgment --- recognizing an undefined metric absent from the refuse-list, and choosing to clarify ambiguity --- are strong yet fallible, and GROUND does occasionally fabricate a plausible metric. Governance should therefore be architected so that the safety-critical guarantees (security, filters, grain) are validated, not merely prompted. This reliability comes at roughly 5$\times$ the token cost of a direct baseline, a gap that prompt caching of the static semantic layer should largely close. Immediate future work is to replicate across models and runs, expand the real-data and adversarial sets, add a query-cost proxy, and study caching\textquotesingle s effect on the cost of governance.

\begin{center}\rule{0.5\linewidth}{0.5pt}\end{center}

\section{Appendix A. Semantic-Layer YAML Excerpt}\label{appendix-a-semantic-layer-yaml-excerpt}

The governed definition style used by GROUND. The full benchmark package includes separate files for metrics, dimensions, joins, RLS rules, and the business glossary.

\begin{lstlisting}
metrics:
  service_revenue:
    label: Service Revenue
    base_fact: fact_repair_order_line
    grain: repair_order_line
    expression: "SUM(fact_repair_order_line.customer_pay_labor_amount + fact_repair_order_line.customer_pay_parts_amount)"
    time_column: fact_repair_order.close_date_key
    required_filter_set: service_customer_pay   # CLOSED + RETAIL + POSTED + ACTIVE
    allowed_dimension_group: service            # dealer, region, service_advisor, vehicle_make, month, ...
  repair_order_count:
    grain: repair_order
    # grain guard: rows come from the LINE fact but the count is over ORDERS
    expression: "COUNT(DISTINCT fact_repair_order.repair_order_id)"
    required_filter_set: service_customer_pay

# rls_rules.yaml (per test_user); glossary time-resolution convention
users:
  user_west_manager:
    predicate: { table: dim_dealer, column: region, op: IN, value: [West] }
time_resolution: "Filter time on dim_calendar.date_value ranges only; never label equality."
undefined_metrics:
  customer_lifetime_value: "Not defined in the semantic layer. Do not invent a formula."
\end{lstlisting}

\section{Appendix B. Benchmark Question Schema}\label{appendix-b-benchmark-question-schema}

\begin{longtable}[]{@{}L{0.25\linewidth}L{0.70\linewidth}@{}}
\toprule\noalign{}
Field & Description \\
\midrule\noalign{}
\endhead
\bottomrule\noalign{}
\endlastfoot
\texttt{question\_id} & Stable benchmark identifier (Q001--Q100). \\
\texttt{task\_type} & Category: \texttt{simple\_metric}, \texttt{group\_by}, \texttt{trend}, \texttt{multi\_table}, \texttt{comparison}, \texttt{ambiguous}, \texttt{security}, trap types, etc. \\
\texttt{test\_user} & Synthetic user whose role determines RLS behavior. \\
\texttt{question} & Natural-language analytics request. \\
\texttt{required\_metric} & Approved semantic metric(s), \texttt{;}-separated; blank for ambiguous requests. \\
\texttt{required\_dimension} & Expected grouping dimension(s), \texttt{;}-separated. \\
\texttt{time\_scope} & Expected time filter, e.g. \texttt{last\_month}, \texttt{Q2\ 2025}, \texttt{year\_to\_date}. \\
\texttt{security\_expectation} & Expected RLS scope, e.g. \texttt{restrict\_to\_user\_scope}, \texttt{all\_active\_dealers}, \texttt{dealer\_101\_only}. \\
\texttt{expected\_behavior} & \texttt{generate\_sql} (variants: \texttt{\_count\_distinct}, \texttt{\_with\_careful\_grain}), \texttt{ask\_clarification}, \texttt{reject\_undefined\_metric}, \texttt{reject\_unknown\_table\_or\_metric}, \texttt{reject\_unsupported\_dimension}, \texttt{restrict\_or\_refuse\_scope}, \texttt{return\_data\_only\_or\_request\_diagnostic\_scope}. \\
\end{longtable}

Example seed question: \emph{"Show customer lifetime value by dealer."} Expected behavior: \texttt{reject\_undefined\_metric}, because \texttt{customer\_lifetime\_value} is not defined in the governed semantic layer.

\begin{center}\rule{0.5\linewidth}{0.5pt}\end{center}

\section{Appendix C. Reproducibility and Artifact Availability}\label{appendix-c-reproducibility-and-artifact-availability}

The complete benchmark, semantic layers, gold SQL (both dialects), system runners, and the evaluation harness are released at \textbf{\url{https://github.com/aravindsp/ground-benchmark}}. All results in this paper can be regenerated from that repository; the committed system outputs (\texttt{results/*.jsonl}) also let a reader reproduce the \emph{scoring} --- including the deterministic filter/RLS numbers --- without any API access.

\textbf{Environment.} Python 3.9. Core dependencies: \texttt{pandas}, \texttt{pyyaml}, \texttt{duckdb}, and the provider SDKs \texttt{anthropic}, \texttt{openai}, \texttt{google-genai} (only the runners that call a model need the SDKs and keys; the scorer and self-test do not).

\textbf{Data provenance.} The synthetic automotive CSVs are committed under \texttt{data/csv/} and are the canonical source; they are loaded into \texttt{data/ground\_synthetic.sqlite}. The real-data domain is derived from public U.S. NHTSA vehicle-safety files (\texttt{NHTSA\_Customer\_Complaints/}, U.S.-Government public domain); \texttt{src/load\_nhtsa.py} projects and types them into \texttt{data/ground\_nhtsa.sqlite} (271,718 complaints) and builds the indexes on which the query-cost proxy depends. NHTSA gold (\texttt{benchmark\_nhtsa/}) is \textbf{hand-authored directly against the database}, independently of the semantic layer.

\textbf{Reproduction.}

\begin{lstlisting}
# 1. Sanity --- no API key needed (scores the gold SQL as an oracle, asserts a clean bill):
for d in synthetic nhtsa nhtsa_hard; do python3 src/evaluate_results.py --self-test --domain $d; done

# 2. Regenerate gold-answer snapshots (synthetic):
python3 src/evaluate_gold_sql_sqlite.py

# 3. Run the four systems (needs ANTHROPIC_API_KEY; OPEN_ROUTER_KEY for the gateway models):
python3 src/run_systems.py --all --domain nhtsa_hard --model claude-opus-4-8
python3 src/run_systems.py --all --domain nhtsa_hard --model "openai/gpt-5.2"   # via gateway

# 4. Score a run, or aggregate run-to-run variance across models:
python3 src/evaluate_results.py --results results/ground.jsonl --label ground
python3 src/score_variance.py --domain nhtsa_hard --models opus,sonnet,gpt52,llama --runs 3
\end{lstlisting}

\textbf{Models and providers.} Four models across three providers were evaluated (runs conducted July 2026): Claude Opus 4.8 (\texttt{claude-opus-4-8}) and Claude Sonnet 5 (\texttt{claude-sonnet-5}) via the Anthropic API; OpenAI GPT-5.2 (\texttt{openai/gpt-5.2}) and open-weight Llama-3.3-70B (\texttt{deepinfra/meta-llama/Llama-3.3-70B-Instruct-Turbo}) via the Requesty gateway (\texttt{https://router.requesty.ai/v1}). Namespaced model ids (containing \texttt{/}) route to the gateway; the runner falls back from strict \texttt{json\_schema} to \texttt{json\_object} to plain-prompt JSON for models that reject structured output.

\textbf{Run configuration.} All systems share one model per comparison and use adaptive thinking with structured JSON output. Only GROUND runs a validate-and-retry loop (default budget 3 retries). GROUND is executed three times per model on the NHTSA standard (Sec. 8.6) and adversarial (Sec. 8.5) sets to report mean $\pm$ standard deviation; the synthetic evaluation (Sec. 8.1) and all baselines are single-run.

\textbf{Determinism note.} GROUND\textquotesingle s \emph{enforced} governance metrics (filter, RLS) are decided by a code-level validation check, not sampled from the model, and reproduce exactly (0.000 $\pm$ 0.000) across runs, models, and machines. The \emph{judgment}-dependent metrics (recognizing an undefined metric, choosing to clarify) are sampled from hosted models and may drift run-to-run and as providers update model snapshots; exact reproduction of those figures is not guaranteed, though the qualitative separation is stable across the four models we tested.

\end{document}